\documentclass{sig-alternate}
\usepackage[ruled,vlined]{algorithm2e}
\usepackage{subfigure}
\usepackage{multirow}
\usepackage{comment, color}
\usepackage{url}

\begin{document}
	%
	
	\title{"Look Ma, No Hands!" A Parameter-Free Topic Model}
	%
	%
	%
	%
	%
	
	\numberofauthors{3} 
	%
	\author{
		%
		%
		\alignauthor
		Jian Tang \thanks{This study is done when the first author is visiting the University of Michigan. }\\
		\affaddr{School of EECS}\\
		\affaddr{Peking University, China}
		\email{tangjian@net.pku.edu.cn}
		\alignauthor
		Ming Zhang\\
		\affaddr{School of EECS}\\
		\affaddr{Peking University, China}
		\email{mzhang@net.pku.edu.cn}
		\alignauthor
		Qiaozhu Mei\\
		\affaddr{School of Information}\\
		\affaddr{University of Michigan, US}
		\email{qmei@umich.edu}
	}
	
	\maketitle
	\begin{abstract}
		It has always been a burden to the users of statistical topic models to predetermine the right number of topics, which is a key parameter of most topic models. 
		Conventionally, automatic selection of this parameter is done through either statistical model selection (e.g., cross-validation, AIC, or BIC) or Bayesian nonparametric models (e.g., hierarchical Dirichlet process). These methods either rely on repeated runs of the inference algorithm to search through a large range of parameter values which does not suit the mining of big data, or replace this parameter with alternative parameters that are less intuitive and still hard to be determined.
		
		In this paper, we explore to ``eliminate'' this parameter from a new perspective. We first present a nonparametric treatment of the PLSA model named nonparametric probabilistic latent semantic analysis (nPLSA). The inference procedure of nPLSA allows for the exploration and comparison of different numbers of topics within a single execution, yet remains as simple as that of PLSA. This is achieved by substituting the parameter of the number of topics with an alternative parameter that is the minimal goodness of fit of a document. We show that the new parameter can be further eliminated by two parameter-free treatments: either by monitoring the diversity among the discovered topics or by a weak supervision from users in the form of an exemplar topic.  The parameter-free topic model finds the appropriate number of topics when the diversity among the discovered topics is maximized, or when the granularity of the discovered topics matches the exemplar topic.
		Experiments on both synthetic and real data prove that the parameter-free topic model extracts topics with a comparable quality comparing to classical topic models with ``manual transmission.'' The quality of the topics outperforms those extracted through classical Bayesian nonparametric models.
	\end{abstract}
	\category{H.3.3}{Information Search and Retrieval}{Text Mining}
	\terms{Algorithm, Experimentation}
	\vskip-5em
	\keywords{Nonparametric models, diversity, parameter-free}

\section{Introduction}
\noindent \textit{How many topics are there in science? 100? 1,000? How many topics are there in Wikipedia? Twitter? The Web?}

Statistical topic models (e.g., \cite{Hofmann:1999:PLS:312624.312649, Blei:2003:LDA:944919.944937}) are widely adopted to analyze text collections in various domains, such as the Web, scientific literature, social media, and digital humanities. 
Because of the principled mathematical foundation and effectiveness in exploratory data analysis, topic modeling has found its way into many classical and new data mining tasks, including topic extraction~\cite{Blei:2003:LDA:944919.944937,griffiths04finding}, sentiment analysis~\cite{DBLP:conf/cikm/LinH09}, link prediction~\cite{Liu:2009:TLJ:1553374.1553460,DBLP:journals/jmlr/ChangB09,Parimi:2011:PFL:2022850.2022857} and collaborative filtering~\cite{DBLP:conf/kdd/WangB11}. 

The success of topic models largely ascribes to its ease of use and interpretability of results in exploratory data analysis. However, it has always been a burden for the users of topic models to predefine the number of topics, which is the key parameter of most topic models. The situation becomes worse when the data is large and when the domain is open, as it is even hard to determine an appropriate range of the number of topics with limited prior knowledge. As a result, the user has to execute the same algorithm many times in order to enumerate different numbers of topics. The optimal number of topics is selected either by manually assessing the quality of topics extracted or through cross validation. Such a ``manual transmission'' process is apparently undesirable when the size of data and the number of topics are large. 

Efforts in literature have attempted to solve this problem through Bayesian nonparametric models. Such models are built on an infinite-dimensional parameter space, thus have the capability to model infinite mixtures of topics and adapt the number of topics to the complexity of the data. 
These nonparametric models typically substitute one parameter (i.e., number of topics) with another (e.g., the concentration parameter in hierarchical Dirichlet process), which is even less intuitive and still hard to tune. These methods also usually rely on a more sophisticated inference algorithm. 
The quality of the topics extracted by the nonparametric models is often compromised comparing to those discovered through classical, parametric topic models. 
In practice, parametric models such as the probabilistic latent semantic analysis (PLSA)~\cite{Hofmann:1999:PLS:312624.312649} and the latent Dirichlet allocation (LDA)~\cite{Blei:2003:LDA:944919.944937} are still widely preferred.

It is our desire to find a user-friendly model that is simple, efficient, and free of parameters. Ideally, such a model should employ an inference procedure as simple as those of the classical models, and thus could be easily implemented and scaled up to handle large scale datasets. The algorithm should either automatically find the appropriate number of topics, or allow a user to interactively guide it towards the desirable number of topics, instead of shooting in the dark.

In this paper, we seek for such a model by exploring a new direction. 
We start with the intuition of how human labels books into categories. Without prior knowledge about the number of categories, a human cataloger would scan the books in order and start with a minimal number of labels. A book will be tagged with a label that it fits in, or a new label will be created for that book and passed along, if no existing label describes the book well. With this strategy, the cataloger does not have to go over all the books again and again. Can a topic model work as smartly as the human cataloger? If the existing topics cannot describe a document, it implies that another topic may exist that is significantly different from the existing ones, thus a new topic should be generated. Embedding such a process into classical topic models, we are able to grow the number of topics without the overhead of separate runs of the inference process. 

To do this, the model has to read from the cataloger's mind for a measure of the ``goodness of fit.'' Inspired by the likelihood ratio test~\cite{citeulike:6095696}, we can measure how well a document is described by the current topics using the likelihood ratio of it being generated by the existing topics versus being generated by all plus one more topic. In this way, the topic model can find the right number of topics whenever a minimal threshold of the goodness of fit is provided. 

Such a treatment actually substitutes the old burden of finding the number of topics with a new burden of finding this minimal threshold of the goodness of fit. We would further eliminate this burden to the users by finding this threshold automatically. One intuition of ours is that a cataloger wants more than the goodness of fit, who also expects the categories be clearly distinguishable. With this intuition, we propose an adaptive process so that a topic model can explore decreasing thresholds and stop when the \textit{diversity} among topics is maximized. We find that the diversity of topics is a good indicator of the optimal number of topics. With such an adaptive process, we present a topic model that finds the appropriate number of topics without arbitrarily defined parameters. 

Another intuition is that a user should be allowed to guide the model towards the number of topics that best fits his personal need. While telling how many topics I anticipate is hard, naming what kind of topics I expect is much easier.  
With a minimal effort, the user can provide an exemplar topic as simple as a keyword. The topic model would then extract topics at a similar granularity to this example. For instance, when a user queries for ``machine learning,'' the model returns topics including ``machine learning,'' ``information retrieval,'' and ``data mining,'' without going deeper, for example, into ``frequent pattern mining.''

To summarize, we present the following contributions:
\begin{enumerate}
  \item We propose a nonparametric version of PLSA, named nPLSA, which dynamically grows the number of topics in a single inference process. 
  Such a treatment substitutes the parameter of the number of topics with a threshold of the minimal goodness of fit, which allows us to further eliminate the parameters.
  \item We propose a parameter-free treatment of nPLSA based on the diversity among the extracted topics, which is shown to be a good indicator of the optimal number of topics. The parameter-free nPLSA finds the appropriate number of topics when the extracted topics are the most distinctive from each other. 
  \item  We propose an alternative parameter-free treatment of nPLSA by receiving weak supervision from users. The model extracts topics at the same granularity of a user-provided exemplar topic (e.g., a keyword). 
  Both models require no arbitrary parameter from the user. 
\end{enumerate}

	
\section{Related Work}
To the best of our knowledge, the two parameter-free treatments of nPLSA are the first topic models that are truly free of any parameters to be predetermined by the users. Traditional clustering algorithms such as k-means~\cite{Hartigan/Wong:79} and spectral clustering~\cite{zha:spectral}, and statistical topic models, such as PLSA and LDA, all require users to specify the number of clusters or topics. One way to choose this parameter is to run the same algorithm on the same data many times with different parameter values, and then choose the best model based on statistical model selection techniques such as the perplexity on held-out data, the Akaike information criterion (AIC)~\cite{citeulike:6081421}, or the Bayesian information criterion (BIC)~\cite{citeulike:90008}. These criteria generally balance the goodness of fit on the training or test data and the complexity of the model. 
All these methods have to execute the inference algorithm many times to search for the parameter, which is costly.

Another direction leads to Bayesian nonparametric models~\cite{RePEc:bes:jnlasa:v:101:y:2006:p:1566-1581,DBLP:conf/nips/BleiGJT03,DBLP:journals/corr/abs-1206-4665,DBLP:journals/jmlr/HannahBP11}, which have been attracting increasing attention in the machine learning community. These models are able to model infinite mixtures and adapt the number of clusters to the complexity of the data. For clustering problems, the Dirichlet process mixture~\cite{hjort10} model is widely used as the nonparametric prior of the mixture components. In~\cite{RePEc:bes:jnlasa:v:101:y:2006:p:1566-1581}, Teh et al. proposed a Bayesian nonparametric topic model called hierarchical Dirichlet process (HDP). In HDP, each document is modeled with an individual Dirichlet process, and the mixture components are shared across documents by using the same base measure, which itself is distributed according to a Dirichlet process. In~\cite{DBLP:conf/nips/BleiGJT03}, a hierarchical latent Dirichlet allocation (hLDA) model was proposed based on the Dirichlet process to infer topic hierarchies from the data. Although these models are flexible and theoretically sound, they all require more sophisticated inference procedures that are difficult to implement or to scale up. 
Bayesian nonparametric models do not mean ``parameter free.'' In fact, 
these nonparametric models generally replace one parameter (i.e., number of topics) with another (e.g., the concentration parameter in HDP), which is even less intuitive and still has to be predefined.  

Some recent effort has revisited classical clustering algorithms from the Bayesian nonparametric viewpoint. In~\cite{DBLP:journals/corr/abs-1111-0352}, Kulis and Jordan designed the DP-means algorithm, which is a scalable clustering algorithm built on top of k-means and is able to infer the number of clusters from the data. DP-means shares a similar iterative process with k-means except that it allows new clusters to be generated during the learning process. The central idea of DP-means is that when the minimum distance between an instance and existing cluster centers is above some threshold, the algorithm will generate a new cluster centered in this instance. The nonparametric PLSA (nPLSA) we will present shares a similar intuition. 
However, the DP-means algorithm still requires users to specify the threshold of the minimal distance, which is even harder than defining the number of clusters. The two parameter-free treatments of nPLSA eliminate arbitrary parameters like such, which can also be applied to the context of data clustering.

	\section{PLSA}
The classical parametric topic model probabilistic latent semantic analysis (PLSA) assumes that each document is a mixture proportion of topics, with each topic corresponding to a multinomial distribution over the word vocabulary. The log-likelihood $L(D)$ of the training data $D$ is calculated by:
\begin{equation}
\label{plsa:likelihood}
  L(D)=\sum_d\sum_w n(d,w)\log\sum_z p(z|d)p(w|z).
\end{equation}
where $n(d,w)$ is the frequency of word $w$ in document $d$, $p(z|d)$ is the probability of topic $z$ in document $d$, and $p(w|z)$ is the probability of word $w$ being generated by topic $z$.
Following the maximum-likelihood principle, the parameters of the model, i.e. $p(z|d)$ and $p(w|z)$,  are estimated by maximizing the log-likelihood objective function (1). Due to the non-convexity of the log-likelihood defined by (1), it is difficult to obtain the global optimum value. To this end, the EM algorithm is generally applied for the problem. The EM algorithm alternates between two steps: E-step and M-step. In the E-step, it calculates the posterior distribution of the latent variable $z$ conditioned on the observation and current model parameters. In the M-step, it updates the model parameters based on the posterior probability calculated in the E-step. We summarize the updating equations as below:

\textbf{E-step:} computing the posterior of the latent variable $p(z|d,w)$ as follows:
\begin{equation}
\label{eqn:estep}
  p(z|d,w)=\frac{p(z|d)p(w|z)}{\sum_{z'} p(z'|d)p(w|z')}
\end{equation}
	
    \textbf{M-step:} updating the model parameters $p(z|d)$ and $p(w|z)$ based on the E-step:
\begin{equation}
\label{eqn:mstep-1}
  p(z|d)\propto \sum_w n(d,w)p(z|d,w)
\end{equation}
\begin{equation}
\label{eqn:mstep-2}
  p(w|z) \propto \sum_d n(d,w)p(z|d,w)
\end{equation}

To apply PLSA for topic modeling, users are required to specify the number of topics, $K$. However, it is generally very hard for users to estimate the number of topics in their datasets a priori, and this brings a considerable burden to the users. In the next section, we introduce a series of methods to eliminate the parameter $K$. 

\section{Eliminating the Parameter $K$}
To eliminate the parameter $K$, or to automatically choose the optimum $K$, one natural way is to execute the PLSA algorithm with different values of $K$ and then choose the value that maximizes the likelihood of a held-out dataset or a statistical model selection criterion, such as AIC and BIC. These methods inevitably require many runs of the algorithm and hence are computationally expensive. A better way is to explore different numbers of topics within a single execution of the algorithm.  Motivated by this, in Section 4.1 we introduce a nonparametric variation of PLSA named nonparametric PLSA (nPLSA), which allows the number of topics to dynamically increase and hence can compare different values of $K$ within a single run of the algorithm. Although the nPLSA model still relies on a user parameter different from $K$, it can be eliminated either by making use of the diversity among the topics (Section 4.2) or weak supervision from users by providing an exemplar topic, which is as simple as simple as a query (Section 4.3). 

\subsection{Nonparametric PLSA}
The nonparametric PLSA (nPLSA) model is motivated by the intuition that if a document is not well fitted by the current topics, then it implies at least one different topic exists in the document, which is not covered by existing topics. One natural reaction is to allow the document to self-promote into a new topic. In this way, this document should fit well into this new set of topics (the union of the original topics and the new topic emitted by the document). This procedure can be naturally applied to a collection of documents in a sequential manner. 

One important issue is how to measure the goodness of fit of a document by the existing topics. Let $\Theta$ be a set of topics and $\theta_d$ be the  language model of document $d$. If a new topic was promoted by the document $d$, an expanded set of topics would emerge as $\Theta'=\Theta \cup \theta_d$. Naturally, the new topic would emerge as the language model of the document $d$. 
This decision can be made by comparing how well the document $d$ is fitted by the two models $\Theta$ and $\Theta'$. If $\Theta'$ gives a much better explanation of $d$ than $\Theta$, then the new topic should be generated; otherwise the current set of topics should be carried on. Formally, a likelihood ratio test~\cite{citeulike:6095696} can be used to compare the goodness of fit of two models, with the requirement that one is nested within the other (in our case, $\Theta$ is nested within $\Theta'$). Inspired by this, we define a distance metric between a document $d$ and a set of topics $\Theta$ as the log-likelihood ratio of the two models $\Theta, \Theta'$:
\begin{equation}
\label{eqn:delta1}
  \Delta(d, \Theta)=\log p(\textbf{w}_d|\Theta'=\Theta \cup \theta_d)-\log p(\textbf{w}_d|\Theta)\footnote{For clarity purpose, we use the notation
   $\log p(\textbf{w}_d|\Theta) $ as the maximum log-likelihood of document $\textbf{w}_d$ fitted by $\Theta$.},
\end{equation}
where $\textbf{w}_d$ refers to all the words in $d$. In reality, since the maximum likelihood estimation of document $d$ will yield to its own language model $\theta_d$,  we have $\log p(\textbf{w}_d|\Theta') =\log p(\textbf{w}_d|\theta_d)$. Eqn.~\eqref{eqn:delta1} becomes:
\begin{equation}
\label{eqn:delta}
  \Delta(d, \Theta)=\log p(\textbf{w}_d|\theta_d)-\log p(\textbf{w}_d|\Theta).
\end{equation}

Let $\epsilon$ be a minimal threshold of the distance above (note this threshold parameter $\epsilon$ will be eliminated in the parameter-free treatments in the next two subsections). A new topic will be generated if $\Delta(d,\Theta)$ is above this threshold, and otherwise $\Theta$ will be carried on. 
This process can be easily integrated into the EM algorithm of the original PLSA model, resulting in a nonparametric algorithm summarized into Algorithm 1. More specifically, when making the inference of each document $d$ in the E-step, we first calculate the goodness of the document $d$ by existing topics $\Theta$, i.e. $\Delta(d, \Theta)$. If $\Delta(d, \Theta)>\epsilon$, the algorithm will generate a new topic initialized as $\theta_d$, and all the words in the document $d$ will be initially labeled with the new topic.  Otherwise, the algorithm will not generate a new topic and instead infer the latent variables $z$ using the current model. 
Note that when a document is visited in one of the EM iterations, the number of topics may have changed since it was visited in the previous iteration. 
That says, if the number of topics equals the number of topics previously fitted into this document, then the inference of this document can be done based on Eqn.~\eqref{eqn:estep}; otherwise, we will need to fit this document with the new set of topics so that the new topics have a chance to be adapted to the document. This can be achieved by a ``fold-in'' process~\cite{Hofmann:1999:PLS:312624.312649}, which updates the topic distribution of the words in the document, i.e. $p(z|d,w)$, and topic proportions of the document, i.e. $p(z|d)$, by maximizing the likelihood of the document. This process continues until no new topic is generated and the log-likelihood converges. 

\begin{algorithm}[h]
 \SetAlgoLined
 \LinesNumbered
 \SetKwIF{If}{ElseIf}{Else}{if}{then}{else if}{else}{endif} 
 \textbf{Input:} {Training data $D$, $\epsilon$: the threshold of the minimal distance $\Delta(d, \Theta)$. }\;
 \textbf{Output:} {Number of topics $K$, the word distributions of topics $\{p(w|z)\}_{z=1,\ldots, K}$ }\;
 \textbf{initialization:} $K=1$, randomly initialize this topic\; set the number of topics fitted to each document $T_d=1$, for each $d=1,\dots|D|$\;
 \While{ not convergence}{
    \textbf{E-step:} for each document $d$ in $\{1,\ldots,|D|\}$
    \begin{enumerate}
      \item Calculate the log-likelihood ratio $\Delta(d,\Theta)$
      \item 
      	   \lIf{$\Delta(d,\Theta)>\epsilon$}  {\\
	   \begin{enumerate}
                 \item \hspace{2mm}     set $K=K+1$,  $\Theta=\Theta \cup \theta_d$\;
                 \item \hspace{2mm}     $p(z=K|w,d)=1$ and $p(z=j|w,d)=0$ for $j < K$\;
            }
            \end{enumerate}
            \lElseIf {$T_d$=$K$}{\\          
                	 \hspace{4mm}   	make inference based on Eqn.~\eqref{eqn:estep}\;
           	 }
	 	\lElse{\\
                  \hspace{4mm}   conduct a ``fold-in'' process \;
                }

             $T_d=K$\;
	
    \end{enumerate}
    \textbf{M-step:} update model parameters $\Theta$ according to Eqn.~\eqref{eqn:mstep-1} and~\eqref{eqn:mstep-2}.
 }
 \caption{The nPLSA model}
\end{algorithm}

\noindent \textbf{A Theoretical Interpretation} \vspace{1em}

 The procedure of Algorithm 1 is similar to that of original PLSA model except that it allows new topics to be generated in the E-step. Although remaining simple, the algorithm does have a principled theoretical interpretation. Specifically, one can prove that the whole procedure is actually monotonically maximizing the following objective function
\begin{equation}
\label{eqn:obj}
 \sum_d\sum_wn(d,w)\log\sum_{z=1}^K p(z|d)p(w|z)-\epsilon K,
\end{equation}
which is the log-likelihood of the training data penalized by the number of topics. The coefficient of the penalization is essentially the threshold $\epsilon$ in Algorithm 1. In other words, the introduction of one extra topic results in the loss of one minimal threshold of likelihood ratio, $\epsilon$. Clearly, a smaller $\epsilon$ results in a larger number of topics. Such an objective function is in fact closely related to the Akaike information criterion (AIC)~\cite{citeulike:6081421} in statistical model selection and also related to the objective in DP-means~\cite{DBLP:journals/corr/abs-1111-0352}. 

We now prove that Algorithm 1 is maximizing the objective function~\eqref{eqn:obj} in both the E-step and the M-step. In E-step, there are three different cases: (1) $\Delta(d,\Theta)>\epsilon$, based on which one can obtain $\log p(w_d|\Theta')-\epsilon(K+1)>\log p(w_d|\Theta)-\epsilon K$, with the likelihood of all other documents unchanged, thus improving the objective in Eqn.~\eqref{eqn:obj}; (2) $T_d=K$, which boils down to the same inference procedure of the E-step in PLSA; and (3) $T_d<K$, which introduces a ``fold-in'' process that maximizes the likelihood of the current document without touching other documents. 
The M-step is the same as classical PLSA procedure, which maximizes the log-likelihood of data without changing the penalty term, thus also improving the objective. The objective~\eqref{eqn:obj} is upper-bounded as long as there is finite number of topics in the data. By the monotone convergence theorem, the algorithm will finally converge to a local maximum of the objective function. 
    
In summary, the nPLSA model is a nonparametric topic model with an inference procedure much easier than those of the Bayesian nonparametric methods. The nPLSA model replaces the parameter $K$ with a different parameter $\epsilon$. Although $\epsilon$ can be as hard to predetermine as the $K$, the benefit of doing this is that now nPLSA can automatically compare different numbers of topics within a single execution of the algorithm. This makes it possible to further eliminating such an arbitrary parameter, or to find the optimal parameter value automatically. 
In the following, we provide two ways to get rid of the parameter 
$\epsilon$ by utilizing the diversity among the topics (Section 4.2) or requesting weak supervision from users, which is an exemplar topic as simple as a query keyword (Section 4.3). 


\subsection{Parameter-free nPLSA through Topic Diversity}
\label{sec:parameter-free}

\begin{algorithm}[h]
 \SetAlgoLined
  \SetKwIF{If}{ElseIf}{Else}{if}{then}{else if}{else}{endif} 
 \textbf{Input:} {Training data $D$} \;
 \textbf{Output:} {Number of topics $K$, the word distributions of topics $\{p(w|z)\}_{z=1,\ldots, K}$}\;
 \textbf{initialization:} $K=1$, randomly initialize this topic \; 
 \While{ not convergence }{
     Calculate Diversity$(\Theta)$; \\
    \textbf{E-step:}  \\
      	  \lIf {Topic diversity does not achieve the optima} {\\
	  \begin{enumerate}
	  	\item calculate the log-likelihood ratio $\Delta(d,\Theta)$ for each document $d$, and find $d^{*}=\text{argmax}_d \Delta(d,\Theta)$ 
               \item  set $K=K+1$,  $\Theta=\Theta \cup \theta_{d^{*}}$\;
                \item $p(z=K|w,d^{*})=1$ and $p(z=j|w,d^{*})=0$ for $j < K$\;
            	\item for $d\neq d^{*}$, conduct a ``fold-in'' process \;
	\end{enumerate}
            }
            \lElse{\\
                     \hspace{4mm} for each $d$, make inference based on Eqn.~\eqref{eqn:estep}\;
                     }   
    \textbf{M-step:} update model parameters $\Theta$ according to Eqn.~\eqref{eqn:mstep-1} and~\eqref{eqn:mstep-2}.
    
 }
	 
 \caption{The Parameter-free nPLSA model}
\end{algorithm}

The nPLSA model makes it possible to gradually grow the number of topics in a single inference procedure. Intuitively, it is desirable to discover the most significant topics first, and then dive deeper into the finer topics. With nPLSA, one can start with a large $\epsilon$ to generate topics that are largely divergent from existing topics, and then gradually decay $\epsilon$ to distinguish topics that are closer together. As long as $\epsilon$ is monotonically decreasing, one is still maximizing the objective~\eqref{eqn:obj} with the smallest $\epsilon$ explored. The question is when to stop generating new topics, or stop exploring a smaller $\epsilon$.

Our treatment makes use of a measure of diversity among the topics as a criterion to stop generating new topics. Diversity is widely studied as a criteria in ranking~\cite{DBLP:conf/kdd/MeiGR10}, query suggestion~\cite{DBLP:conf/aaai/MaLK10} and document summarization~\cite{Carbonell:1998:UMD:290941.291025}. Some existing work has used diversity as an evaluation criteria for topic quality~\cite{mimno:optimizing}, but not for selecting the optimum number of topics (compare across different numbers of topics). We formally define the diversity among a set of topics as follows:
\begin{equation}
  \text{Diversity}(\Theta)=\frac{2}{(K-1)K}\sum_{i=1}^{K-1}\sum_{j=i+1}^{K}dist(\theta_i,\theta_j), 
\end{equation}
where $K$ is the number of topics, $\theta_i$ is the word distribution of the $i^{th}$ topic, and $dist(\cdot,\cdot)$ is a distance function between the two models. In practice, one can instantiate the distance function with the $L_2$ distance or the Jensen$-$Shannon divergence~\cite{Cover:1991:EIT:129837} (the symmetric version of the Kullback$-$Leibler divergence~\cite{Cover:1991:EIT:129837}). In this work, we simply adopt the $L_2$ distance because it is more sensitive to the top-ranked words. 

Intuitively, if the number of topics is small, the learned topics tend to be close to the background language model and thus do not distinguish well between each other. When the number of topics grows, the granularity of topics becomes finer and the topics become more distinguishable, thus increasing the diversity. However, when the number of topics becomes too large, we start to obtain many small topics which may be too close to each other, which decreases the topic diversity. 
Therefore, diversity seems to be a good measure to capture the right granularity of topics. 

To verify this, we conducted topic modeling analysis on a synthetic dataset, in which the correct number of topics is known (see Figure~\ref{fig:simulation_automatic} in Section~\ref{sec:exp}). The result confirms our intuition and proves 
that topic diversity can be a good indicator of the right number of topics. 

Therefore, we utilize the diversity among topics as a criteria for stopping the generation of new topics in nPLSA, which yields a new process shown in Algorithm 2. Specifically, we adopt a farthest-first heuristic. In each iteration, we first find the document $d^{*}$ that has the largest distance $\Delta(d, \Theta)$ to existing topics $\Theta$. Then a new topic is generated from document $d^{*}$, i.e. $\Theta=\Theta \cup \theta_{d^{*}}$. This heuristic is related to the process for sampling seedling points in the K-Means++ algorithm~\cite{arthur2007k}. This means exactly one new topic is generated in each EM iteration. We do foresee relaxations of this heuristic so that multiple new topics could be generated in a single iteration, which will further improve the efficiency of the algorithm. For other documents, a ``fold-in'' process is conducted to fit them with the new configuration of topics. During the process, the diversity among the current set of topics is monitored. When the diversity of the topics achieves an optima, the algorithm stops generating new topics and the original inference procedure of PLSA will be conducted until convergence. Such a process does not rely on any predefined parameter, neither $K$ or $\epsilon$.  

The parameter-free nPLSA model through topic diversity provides an automatic way of finding the number of topics that best fits the data. In practice, a user may have a personal preference about the granularity of topics. For example, he or she may want to explore topics at a higher level like  ``data mining'' or ``machine learning,'' while another user may explore topics with a finer granularity like ``frequent pattern mining'' or ``semi-supervised learning.'' A friendly topic model may adapt the number of topics to this personal need. Below we present another parameter-free treatment to nPLSA, which utilizes weak supervision from a user and finds the number of topics that best fits the need of the user. 

\subsection{Parameter-free nPLSA through Weak Supervision}
The alternative parameter-free treatment of nPLSA allows a user to steer the topic modeling process by providing limited supervision. In practice, we found that while it is hard for a user to define the number of topics in a given dataset, it is usually much easier for her to describe an exemplar topic of her expectation (e.g., ``\textit{I want to find topics like data mining.}''). To minimize the burden on a user, we thus allow her to provide an exemplar topic in the form of a keyword query, to guide the algorithm towards the desirable granularity of topics. This query-by-example design makes a much more reasonable assumption about the users, because it is an everyday practice to construct search queries. 

Let us define the distance of a set of topics $\Theta$ to the query $q$ as the distance between $q$ and the topic closest to it:
\begin{equation}
	d(\theta_q, \Theta)=\min_{\theta_i \in \Theta} L_2(\theta_q, \theta_i),
\end{equation}
where $\theta_q$ is the query language model. While a query can be as short as a keyword, it is the common practice in information retrieval to estimate a robust $\theta_q$ though model-based pseudo feedback~\cite{DBLP:conf/cikm/ZhaiL01}. We first retrieve all the documents containing $q$, which are called feedback documents $F$. Each feedback document is assumed to be a mixture of the query model $\theta_q$ and a collection background model $\theta_C$. $\theta_q$ is estimated by maximizing the likelihood of the feedback documents $F$ under this mixture assumption. 

In the process of Algorithm 2, the granularity of topics decreases as they are split into subtopics. 
Initially the algorithm may identify a topic with a coarse granularity that is closest to $\theta_q$. As the algorithm runs, this topic is split into subtopics, and one of its subtopics may become closer to $\theta_q$. 
During this process, $d(\theta_q,\Theta)$ will decrease. When the granularity of the topics matches the granularity of $\theta_q$, the distance $d(\theta_q, \Theta)$ will achieve the minimum. Now when the topic closest to $q$ is further split into smaller topics, the centers of the subtopics are likely to move away from the center of $\theta_q$ (see our empirical results in Figure~\ref{fig:querybyexample}). The algorithm thus stops generating new topics when $d(\theta_q,\Theta)$ achieves a minimum, and at that point it discovers topics with the same granularity as $q$. The whole process remains the same as Algorithm 2 except that the criterion for stopping generating new topics becomes minimizing $d(\theta_q, \Theta)$ instead of maximizing diversity. 

To summarize, the two parameter-free treatments of nPLSA automatically find the number of topics that are either the most distinctive from each other or the best fits to the user's need, without enumerative runs of the inference algorithm. Next we provide more discussion on some practical issues of the above three models. 

\subsection{Discussion}
\noindent \textbf{How to determine when topic diversity is maximized?}
Since the diversity among the topics may not be a convex function with respect to the number of topics, in practice we can wait several iterations until the topic diversity begins to decrease smoothly. An alternative solution is to run the parameter-free nPLSA with enough iterations to see how the topic diversity changes, from which we can figure out the optimum number of topics. Then one can simply execute a classical topic model using this optimum number of topics. It is similar to determine when $d(\theta_q, \Theta)$ achieves the minimum. 

\noindent \textbf{Sensitivity to the order of documents.} 
As the nPLSA model introduced in Section 4.1 processes the documents in sequential order,  one concern is whether the order of the documents affects the result. We will empirically show that the result is not sensitive to the order of the documents (See Figure~\ref{fig:sensitivity_dblp} in Section 5).  The parameter-free treatment of nPLSA with either topic diversity or weak supervision does not depend on the order of the documents. In each iteration, each of the two scans the entire dataset to get the farthest document and generate a new topic from it; for the rest of the documents, a ``fold-in'' process is conducted to fit them with the current set of topics.  

\noindent \textbf{Scalability.}
The parameter-free nPLSA introduced in Section 4.2 and 4.3 are batch algorithms. To suit the mining of streaming and big data, extending them into online inference algorithms appears to be more reasonable than making them parallel. 
We leave the online inference procedure of parameter-free nPLSA for future work.

	\section{Experiments}
\label{sec:exp}
We evaluate the nPLSA model and its parameter-free versions against classical topic models PLSA and LDA, and also the Bayesian nonparametric topic model HDP\footnote{We adopt the implementation of HDP from \url{http://www.cs.cmu.edu/~chongw/resource.html}. The concentration parameters for both levels are set as 1 by default.}. We evaluate all the models on both synthetic and real-world datasets.

\subsection{Evaluation Metrics}
We first introduce the evaluation metrics. On synthetic data, we know the ground truth of the topics. We denote the ground truth topics as $\Theta^{*}=\{\theta_k^{*}\}_{k=1,\ldots K^{*}}$, and the learned topics as $\Theta=\{\theta_j\}_{j=1,\ldots K}$, where $K^{*}, K$ are the number of topics in ground truth and the number of topics learned by the algorithm(s). The quality of the learned topics can be measured by comparing them to the ground-truth topics.

\vspace{1em}\noindent \textbf{Topic Quality Error.} The topic quality error measures the error that the learned topics introduce to the ground truth topics. The quality error of each individual learned topic is defined as the minimum distance between the topic and the ground truth topics. The overall topic quality error is averaged over all the learned topics:
\begin{equation}
\small
  \text{TQE}=\frac{1}{K}\sum_{j=1}^{K}\min_{k=1,\ldots K^{*}}L_2(\theta_j, \theta_k^{*}).
\end{equation}
On real data, the ground truth is unknown. We introduce two metrics that do not depend on the ground truth of the topics. One of them measures the semantic coherence of the topics discovered~\cite{DBLP:conf/jcdl/NewmanNTKB10,newman2010automatic,DBLP:conf/emnlp/MimnoWTLM11}. In~\cite{DBLP:conf/jcdl/NewmanNTKB10}, the point-wise mutual information is used.  Finally, we present the metric of perplexity, which is a commonly used metric in literature.

\vspace{1em}\noindent \textbf{Topic Semantic Coherence.} The topic semantic coherence is used to measure the semantic coherence of the learned topics. For each topic, the semantic coherence is defined as the average point-wise mutual information score of every pair of the top-ranked words in a topic $\theta \in \Theta$:
\begin{equation}
\small
  \text{PMI}(\Theta)=\frac{1}{K}\sum_{\theta \in \Theta}\frac{2}{N(N-1)}\sum_{1 \leq i<j \leq N}\log \frac{p(w_{i, \theta},w_{j, \theta})}{p(w_{i, \theta})p(w_{j, \theta})},
\end{equation}
where $w_{i, \theta}$, $w_{j, \theta}$ are the words ranked at the $i^{th}$ and $j^{th}$ positions in topic $\theta$. $p(w_{i, \theta}, w_{j, \theta})$ is the probability that the pair of words co-occur in the same document while $p(w_{i, \theta})$ is probability of a word $w_{i, \theta}$ appearing in a single document. In our experiments, we focus on the top 20 words per topic ($N$ = 20). A large dataset is required to calculate the point-wise mutual information of the word pairs. 

\vspace{1em}\noindent \textbf{Perplexity.} The perplexity is used to measure the predictive performance of the topics learned on a held-out dataset. Specifically, for each document $\textbf{w}_j$ in the held-out dataset, we split the document into two parts $\textbf{w}_j=(\textbf{w}_{j1}, \textbf{w}_{j2})$ and compute the predictive likelihood of the second part $w_{j2}$ ($20\%$ of the words) conditioned on the first $80\%$ of the words and the training data, the same way used in~\cite{DBLP:conf/icml/WallachMSM09}.  It can be calculated by:
\begin{equation}
\small
  perplexity=\exp\big\{-\frac{\sum_j\log p(\textbf{w}_{j2}|\textbf{w}_{j1}, D_{\text{train}})}{\sum_j{|\textbf{w}_{j2}|}}\big\}
\end{equation}

\subsection{Results on Synthetic Dataset}
 We first compare the behaviors of models on a synthetic dataset so that we know the ground truth of topics. We generate a collection of 1,000 documents with 200 words each using the generative process of LDA with 20 topics, a vocabulary size of 1,000,  and the parameters of Dirichlet priors for topic mixtures and word distributions being 0.1 and 0.01. To make the underlying topics differentiable, 
 we only keep topics whose minimum distance to existing topics is larger than 0.5.

\begin{figure}[htbp]
\centering
\subfigure[Topic quality error (Synthetic) ]{
\includegraphics[width=1.5in]{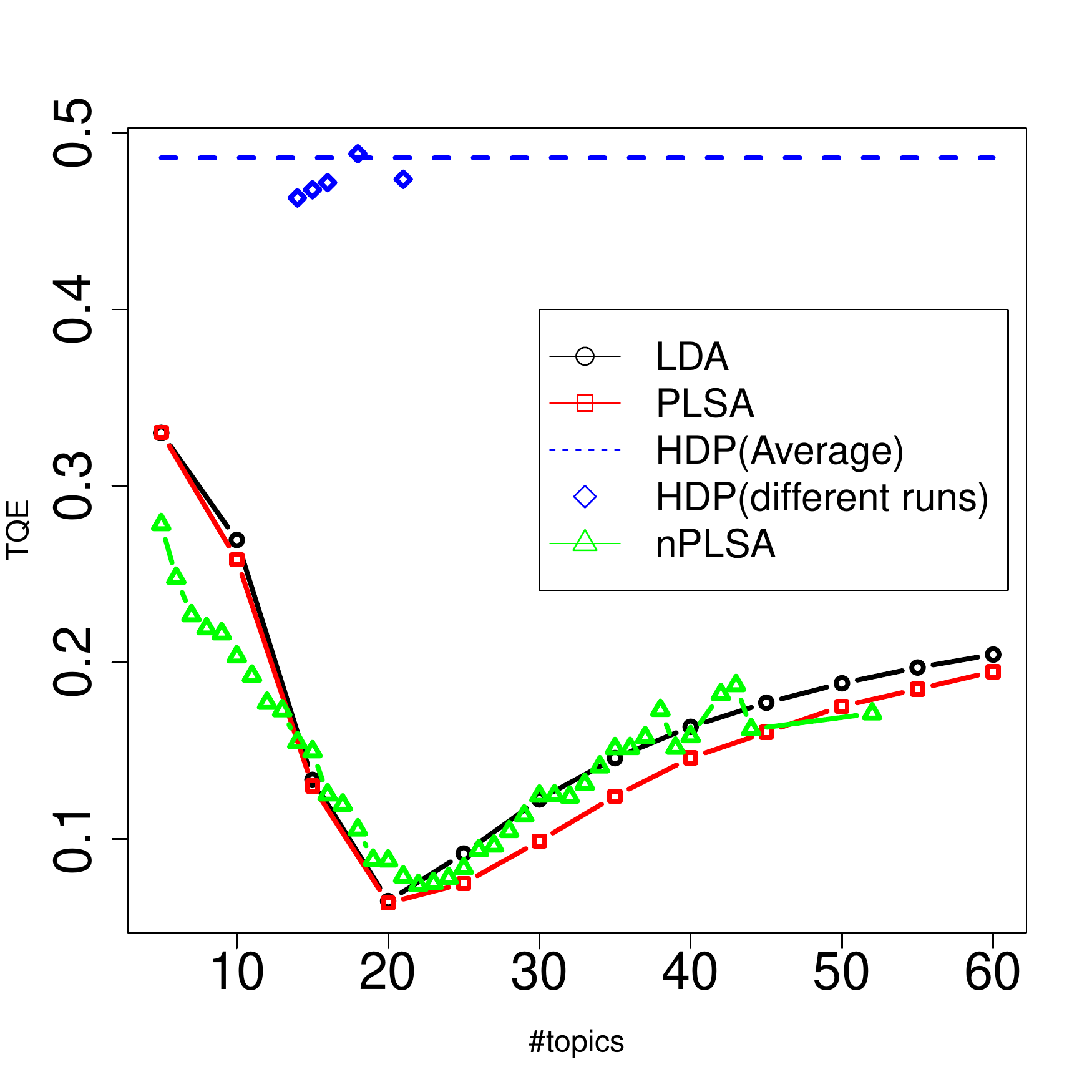}
\label{fig:subfig1}
}
\subfigure[Topic diversity (DBLP)]{
\includegraphics[width=1.5in]{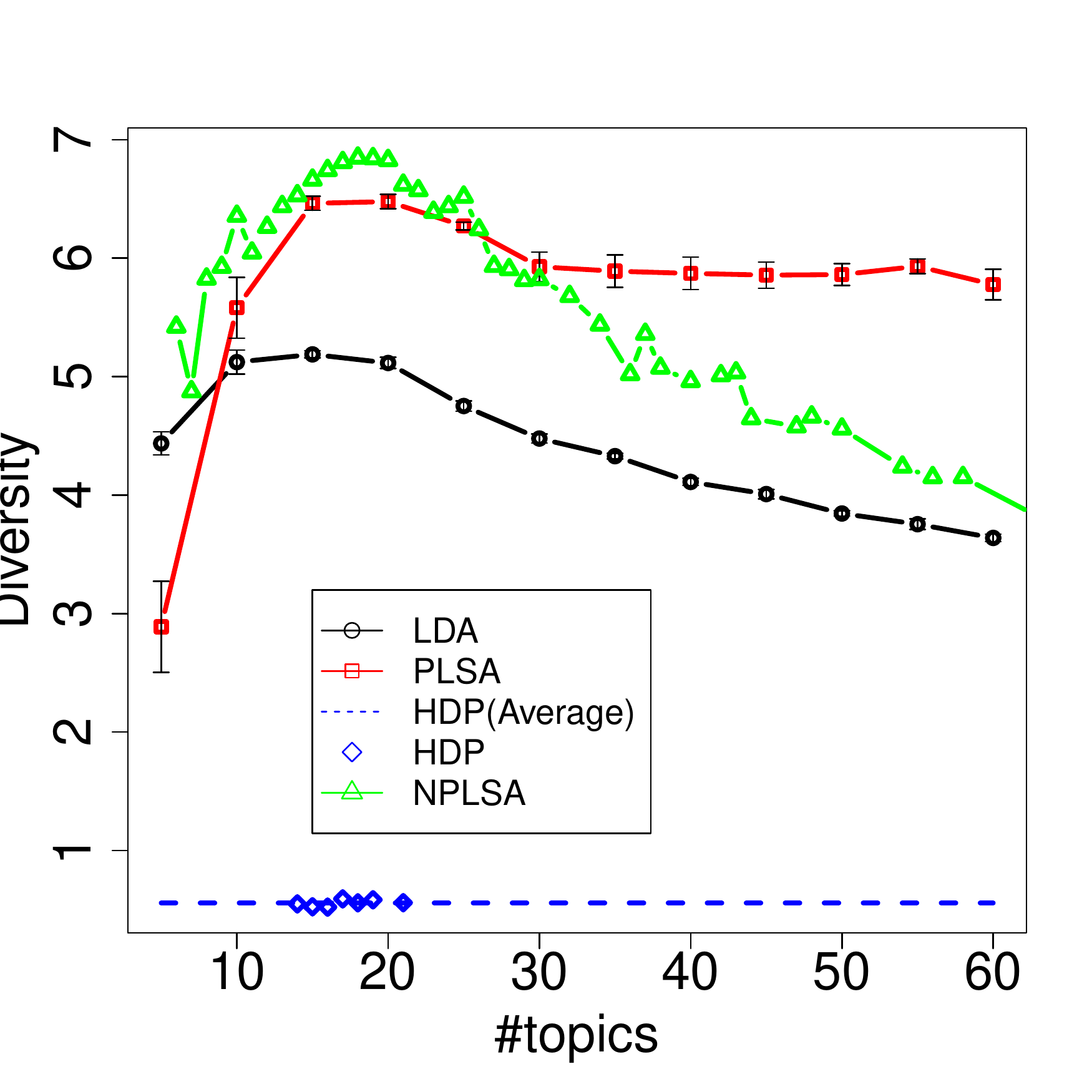}
\label{fig:subfig3}
}
\label{fig:synthetic_result}
\caption{The performance of the nPLSA on both synthetic and real-world datasets. nPLSA achieves comparable results with PLSA and LDA and outperforms HDP. Figure (a): the lower value the better results. Figure (b): the higher value the better results. From (b), we can see that diversity among the topic is a good indicator of the optimum number of topics.}
\end{figure}

Figure 1 presents the results evaluated by different metrics on the synthetic datasets. All the results are averaged over 10 runs with random data generation. Figure 1(a) shows the topic quality errors of different models w.r.t. different numbers of topics. For the nPLSA model, we vary $\epsilon$ from 10 to 400, and plot the results w.r.t the number of topics the model generates. For HDP, we use the default settings and plot the performance similarly (each run corresponds to one data point on the figure ). The performances of PLSA, LDA and nPLSA are comparable and all significantly outperform HDP. The optimum performances of PLSA, LDA, and nPLSA are achieved around the ground truth number of topics. 

The most interesting observation presents in Figure 1(b), which plots the diversity of the topics learned by different models. The diversity of the topics discovered by the nPLSA model is also comparable with that of PLSA and LDA, much higher than that of HDP. Reconfirming our intuition in Section~\ref{sec:parameter-free}, the topic diversity increases when the number of topics increases from a small number of topics, peaks when the number of topics arrives at the ground truth, and drops when even more topics are generated. 
Note that diversity measure does not rely on the ground truth topics, suggesting that it is a good criterion for topic modeling in real data. 

\vspace{1em}\noindent \textbf{Results of the parameter-free nPLSA}\vspace{1em}

Next, we investigate the performance of the parameter-free nPLSA model (Algorithm 2). Figure 2(a) presents the diversity of the topics discovered by the parameter-free nPLSA model in a single run along the iterations or the number of topics (exactly one topic is generated per iteration). The topic diversity monotonically increases to the optima when the number of topics hits 21, and then monotonically drops. This means the parameter-free nPLSA model will yield 21 topics, which is very close to the ground truth, 20. 
Figure 2(b) shows the distributions of $\Delta(d,\Theta)$ of all the documents w.r.t. the number of topics. Overall, the average of $\Delta(d,\Theta)$ is decreasing as more topics are generated. This means that the documents are fitted better with more topics. When the number of topics reaches the ground truth, the distribution of $\Delta(d, \Theta)$ becomes stable. This is because the nPLSA model has already discovered all the underlying topics, and introducing more topics will hardly fit the documents better.
\begin{figure}[htdp]
\centering
\subfigure[Diversity vs \#topics]{
\includegraphics[width=1.5in]{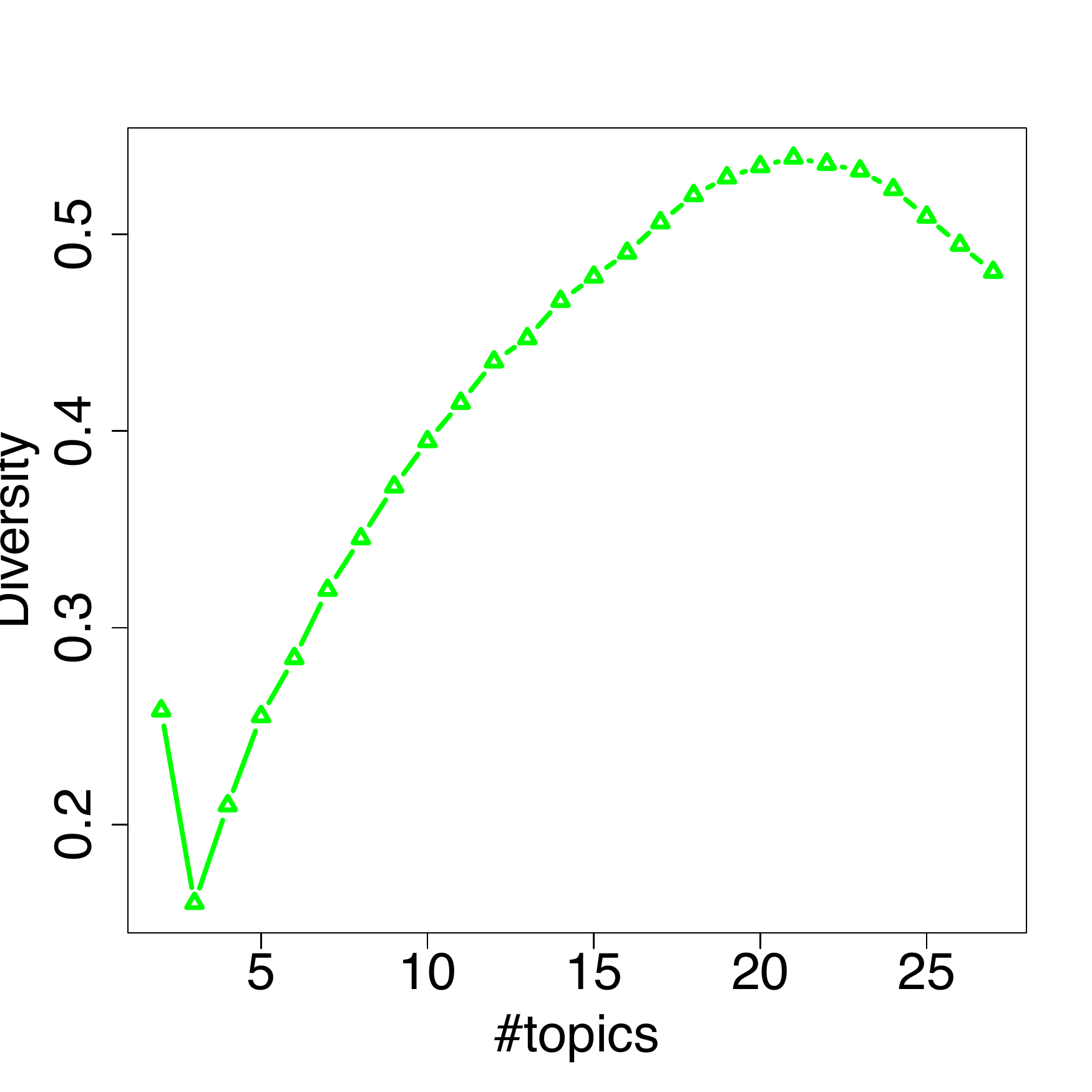}
\label{fig:simulation_automatic}
}
\subfigure[$\Delta(d,\Theta)$ vs \#topics]{
\includegraphics[width=1.5in]{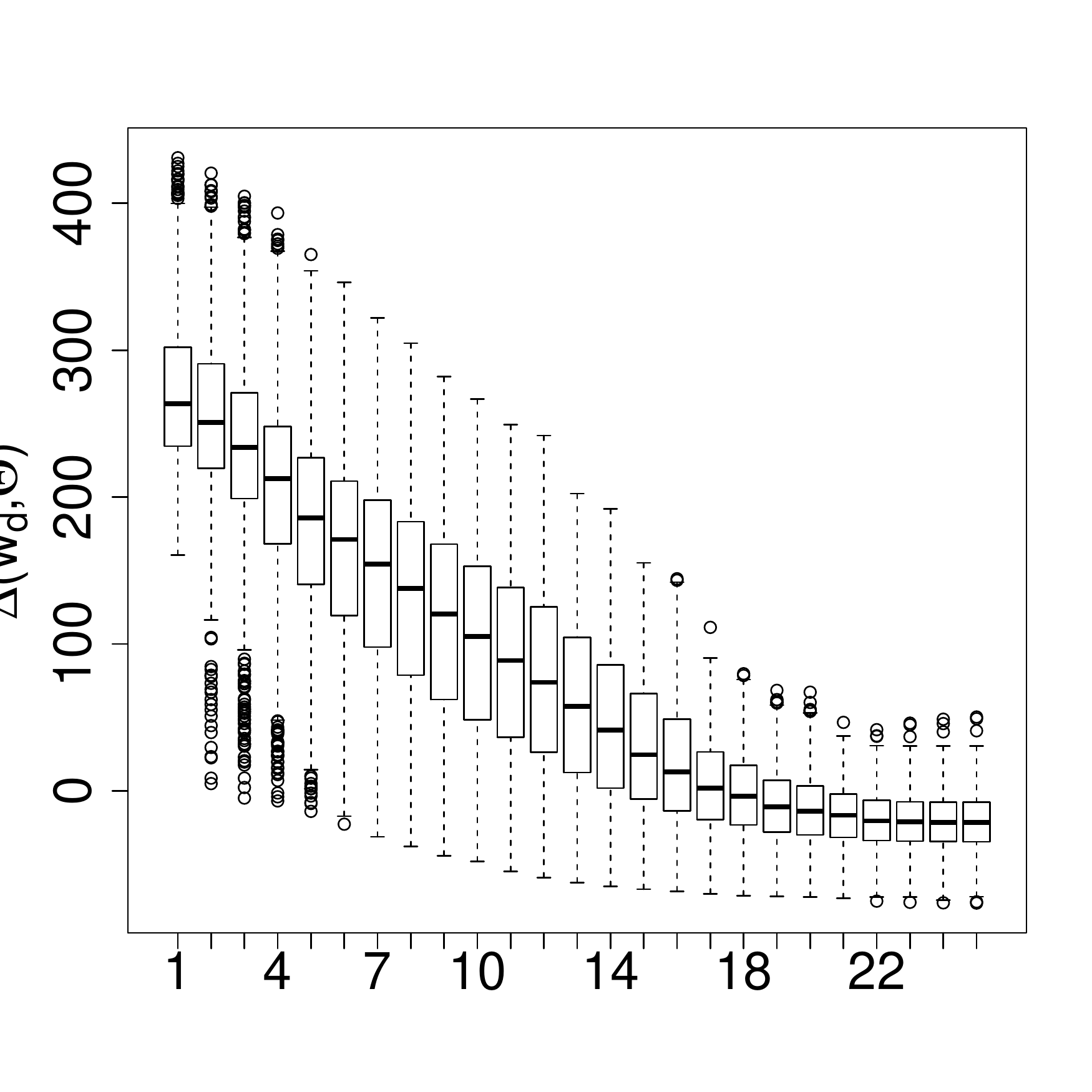}
\label{fig:subfig3}
}
\caption{Results of the parameter-free nPLSA on synthetic dataset in a single run. Figure (a) shows the maximum topic diversity is achieved around the optimum number of topics. Figure (b) shows that the document are fitted better when more topics are generated and this becomes stable once the number of topics generated by the model reaches the ground truth.}
\end{figure}

To summarize, on synthetic data the nPLSA model can obtain comparable performance as PLSA and LDA and significantly outperforms HDP. The diversity of the topics will achieve the optima around the right number of topics, making it a good criteria for selecting the right number of topics. With a single run, the paramter-free nPLSA model can find this optima and yield the appropriate number of topics. 

\subsection{Results on Real-World Datasets}
Besides the promising results on synthetic data, we step forward to evaluate all the models on real-world datasets. The first dataset we use is the computer science bibliography DBLP from~\cite{DBLP:conf/kdd/TangZYLZS08}~\footnote{\url{http://arnetminer.org/citation} }. It includes the records of 1,572,277 papers, from which we kept all the papers with complete abstracts, which yields 529,386 abstracts. We randomly select 5,000 abstracts from the collection and use them as the training data. We removed all the stop words and words that appeared in less than 20 training documents. In this way, 2,899 unique words are obtained. We conducted a 10-fold cross validation on the training dataset. For calculating the semantic coherence of the topics, we utilized the whole abstract dataset.

Figure 3 presents the behaviors of different models on the DBLP dataset. Figure 3(a) plots the perplexity of held-out data against the number of topics. The perplexity of PLSA, LDA and nPLSA all becomes saturated when the number of topics is sufficiently large. The nPLSA performs comparably with PLSA and LDA and outperforms HDP. Note that directly using the perplexity metric as the stopping criterion for nPLSA seems problematic, as it saturates but does not actually converge. 

Figure 3(b) compares the semantic coherence of the topics discovered by different models, measured by the pointwise mutual information. For PLSA, LDA, and nPLSA, the metric presents a clear bell shape, which peaks when the number of topics is moderate (e.g., generally between 50 and 150). The performance of the three models well outperforms HDP. 
Indeed, when reading the actual topics discovered by HDP, we found that it discovered some very smallish topics, which are very difficult to interpret. This behavior may be due to the rich-gets-richer nature of the Dirichlet process. 


\begin{figure}[htbp]
\centering
\subfigure[Perplexity]{
\includegraphics[width=1.5in]{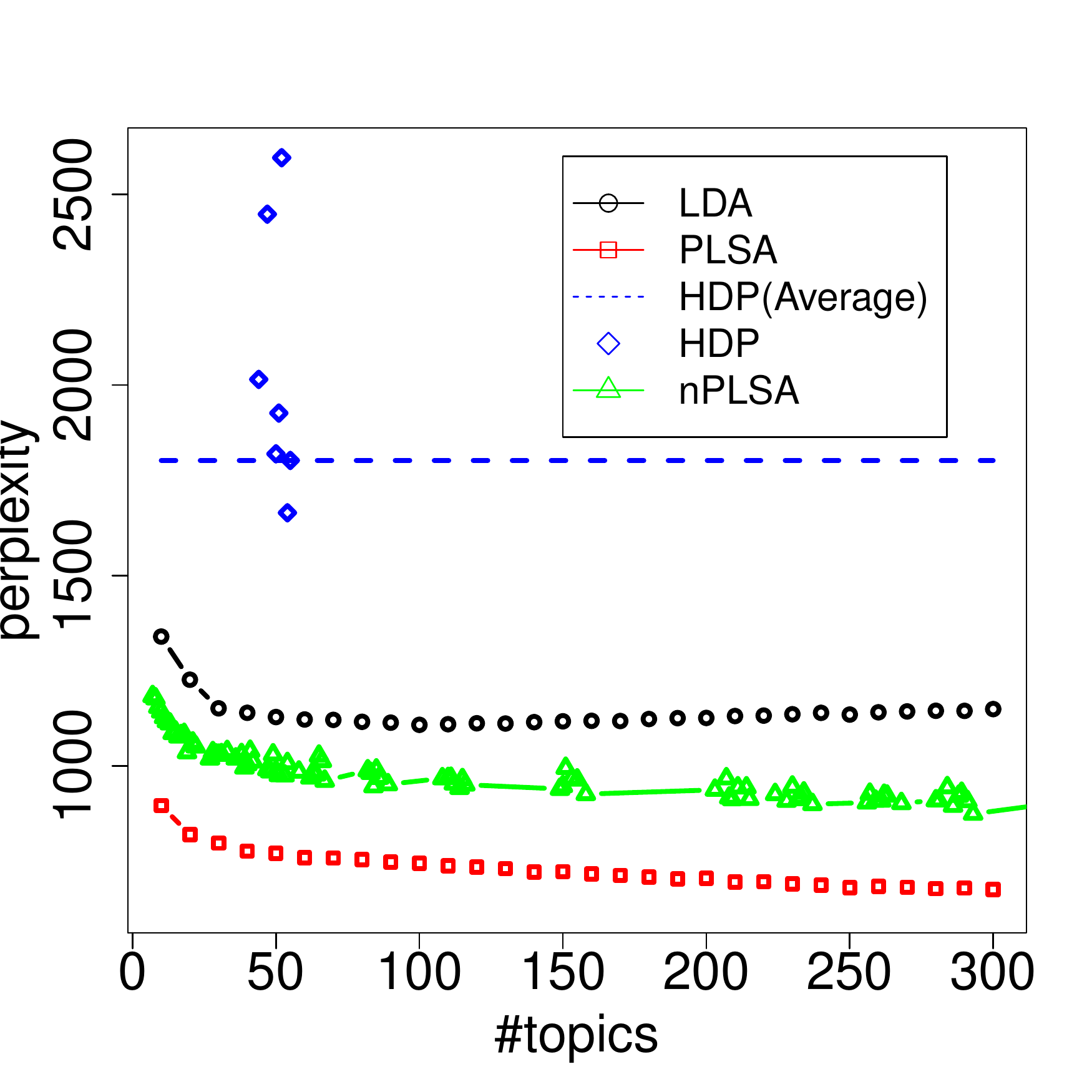}
\label{fig:subfig1}
}
\subfigure[Topic coherence]{
\includegraphics[width=1.5in]{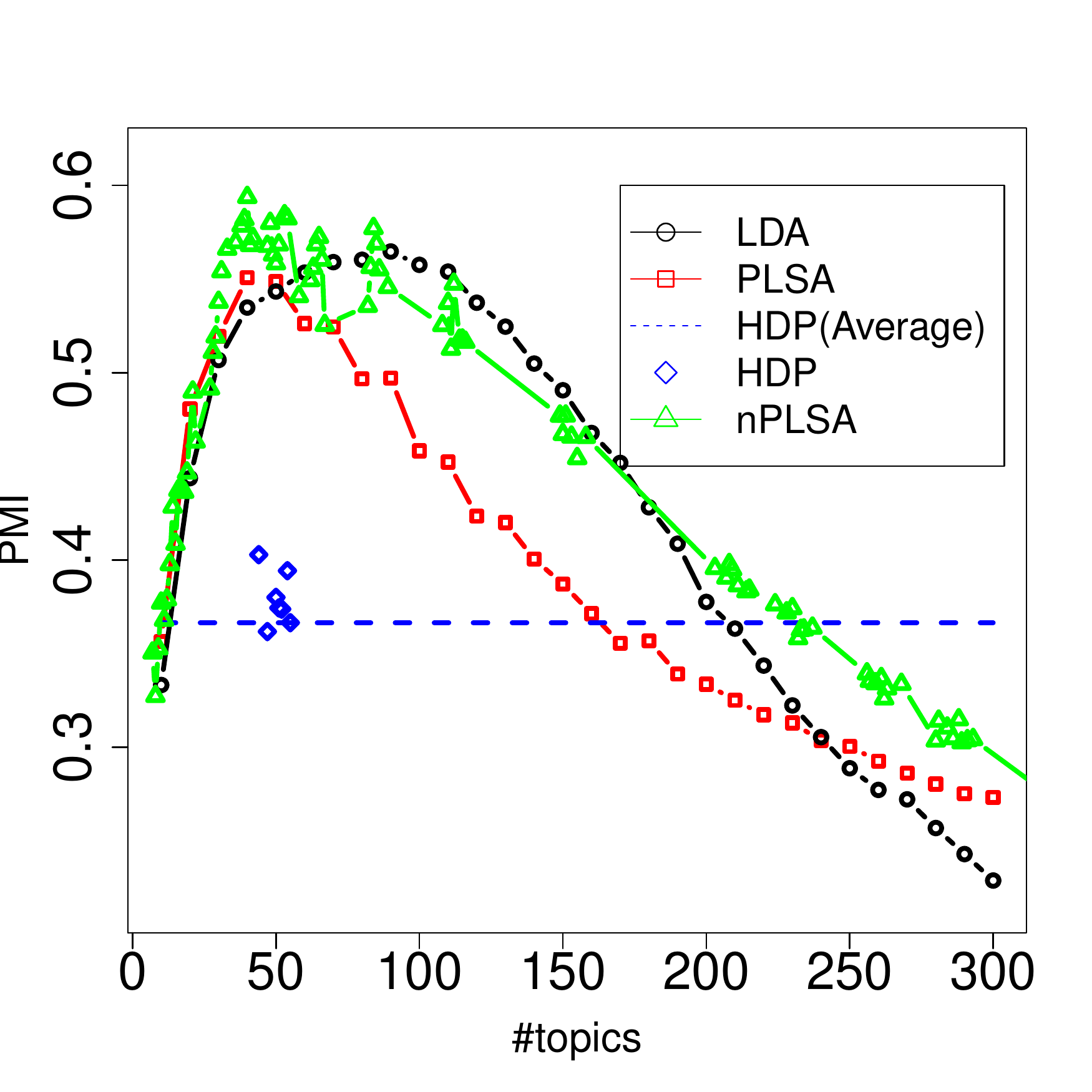}
\label{fig:subfig3}
}
\vskip-1em
\subfigure[Diversity among discovered topics]{
\includegraphics[width=1.5in]{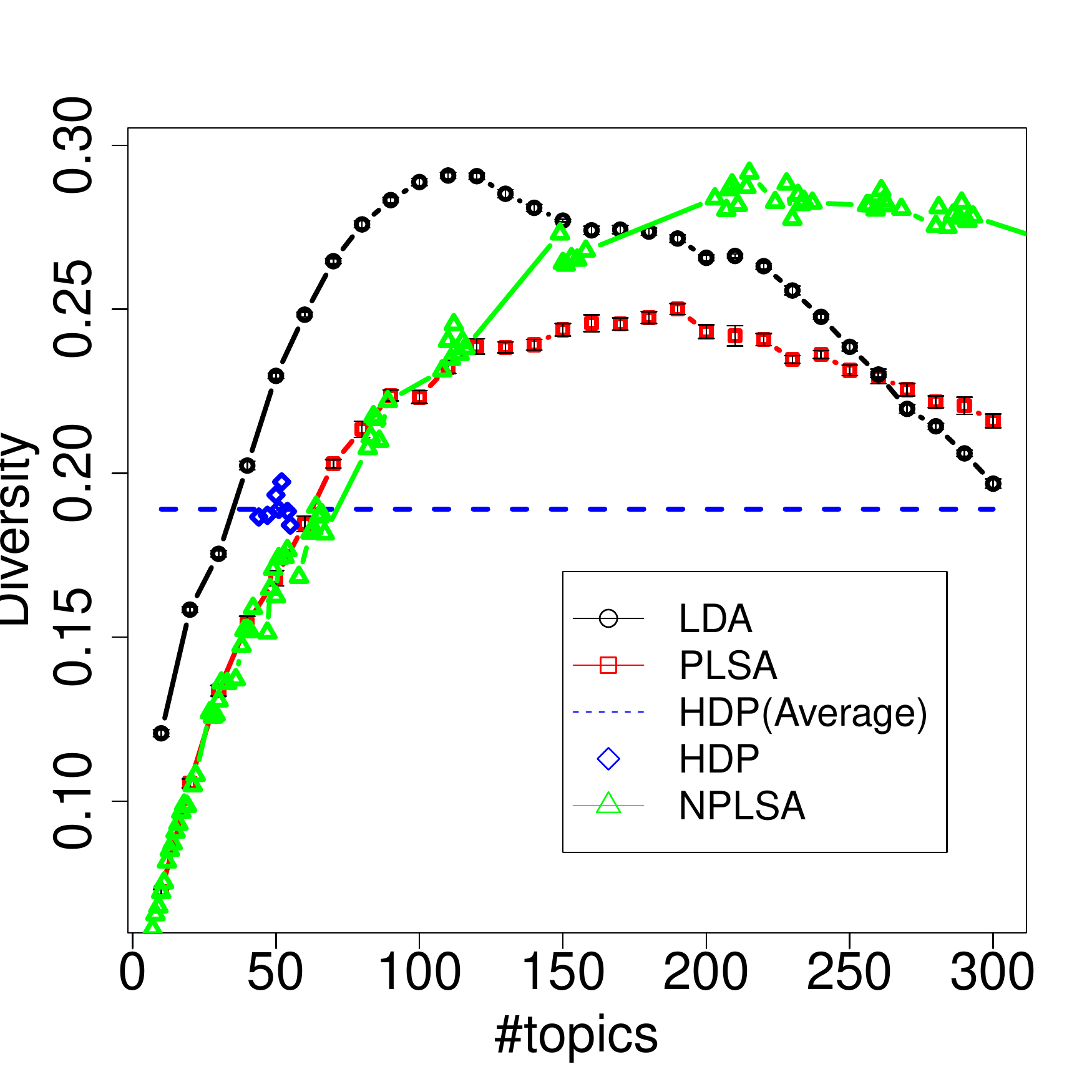}
\label{fig:subfig3}
}
\subfigure[Sensitivity to the order of documents]{
\includegraphics[width=1.5in]{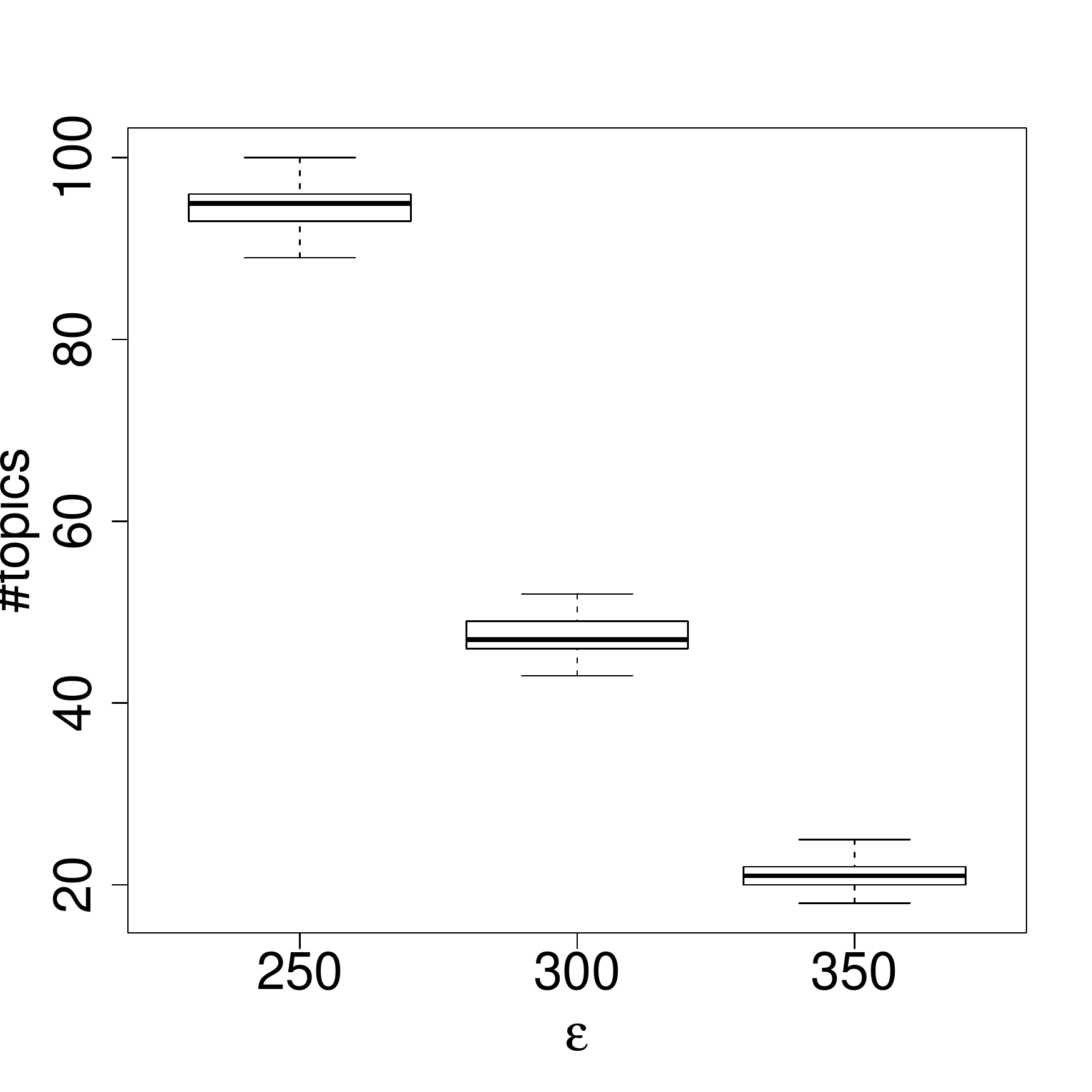}
\label{fig:sensitivity_dblp}
}
\label{fig:subfigureExample}
\caption{Parametric and nonparametric topic models on DBLP dataset. nPLSA achieves comparable results with PLSA and LDA and outperforms HDP. Figure (a): the lower value the better results. Figure (b) and (c): the higher value the better results.}
\end{figure}

Figure 3(c) compares the topic diversity of different models w.r.t different number of topics. The diversity of PLSA, nPLSA, and LDA increases at the beginning and then begins to drop after the number of topics is large enough. This observation is consistent with that on synthetic data. Interestingly, the diversity of topics extracted by nPLSA outperforms the diversity of topics extracted by PLSA (and LDA) when the number of topics is large, given how similar of inference process is. This indicates that the generation process of new topics in nPLSA finds better seedling topics, where in PLSA and LDA all topics are initialized randomly. 

One may wonder how much the process of nPLSA is sensitive to the order in which the documents are processed. In Figure 3(d), we plot the distribution of the number of topics over 50 randomized orderings of the documents. The number of topics discovered by nPLSA is not sensitive to the ordering of documents. 

\vspace{1em}\noindent \textbf{Results of the parameter-free nPLSA on DBLP}\vspace{1em}

Figure 4(a) presents the diversity between the topics generated by the parameter-free nPLSA model (Algorithm 2) in a single run w.r.t the number of topics discovered. Clearly, the diversity of topics increases with more topics at the beginning, reaches an optima around 140, and decreases when more topics are generated. Consistent with the behavior on synthetic data, the parameter-free nPLSA model is able to find the appropriate number of topics by maximizing the diversity metric. Figure 4(b) shows how the parameter $\epsilon$ (the maximum $\Delta(d, \Theta)$ in each iteration) changes w.r.t the number of iterations (topics).

\begin{figure}[htbp]
\centering
\subfigure[Diversity vs $\#$topics]{
\includegraphics[width=1.5in]{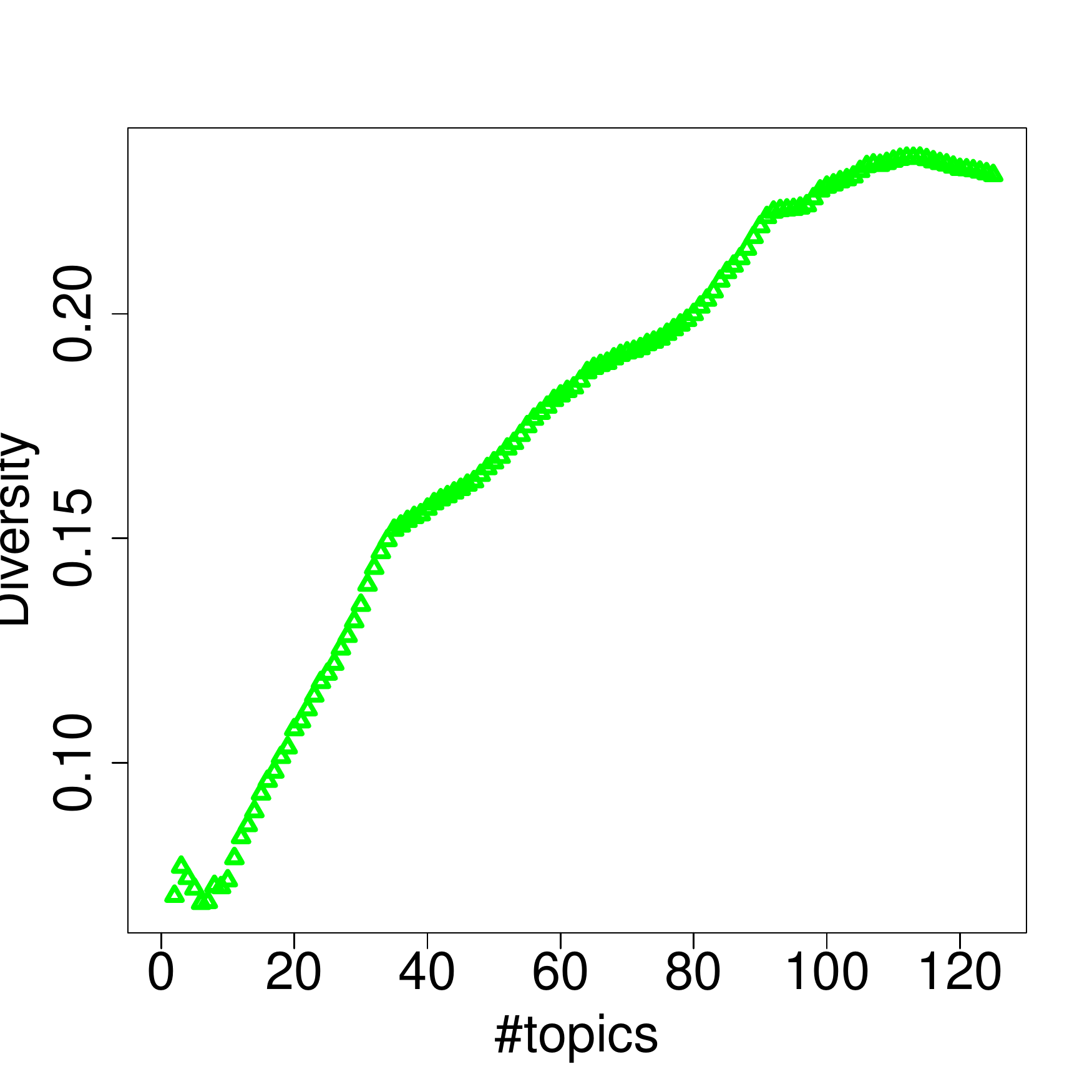}
\label{fig:subfig1}
}
\subfigure[$\epsilon$ vs $\#$topics]{
\includegraphics[width=1.5in]{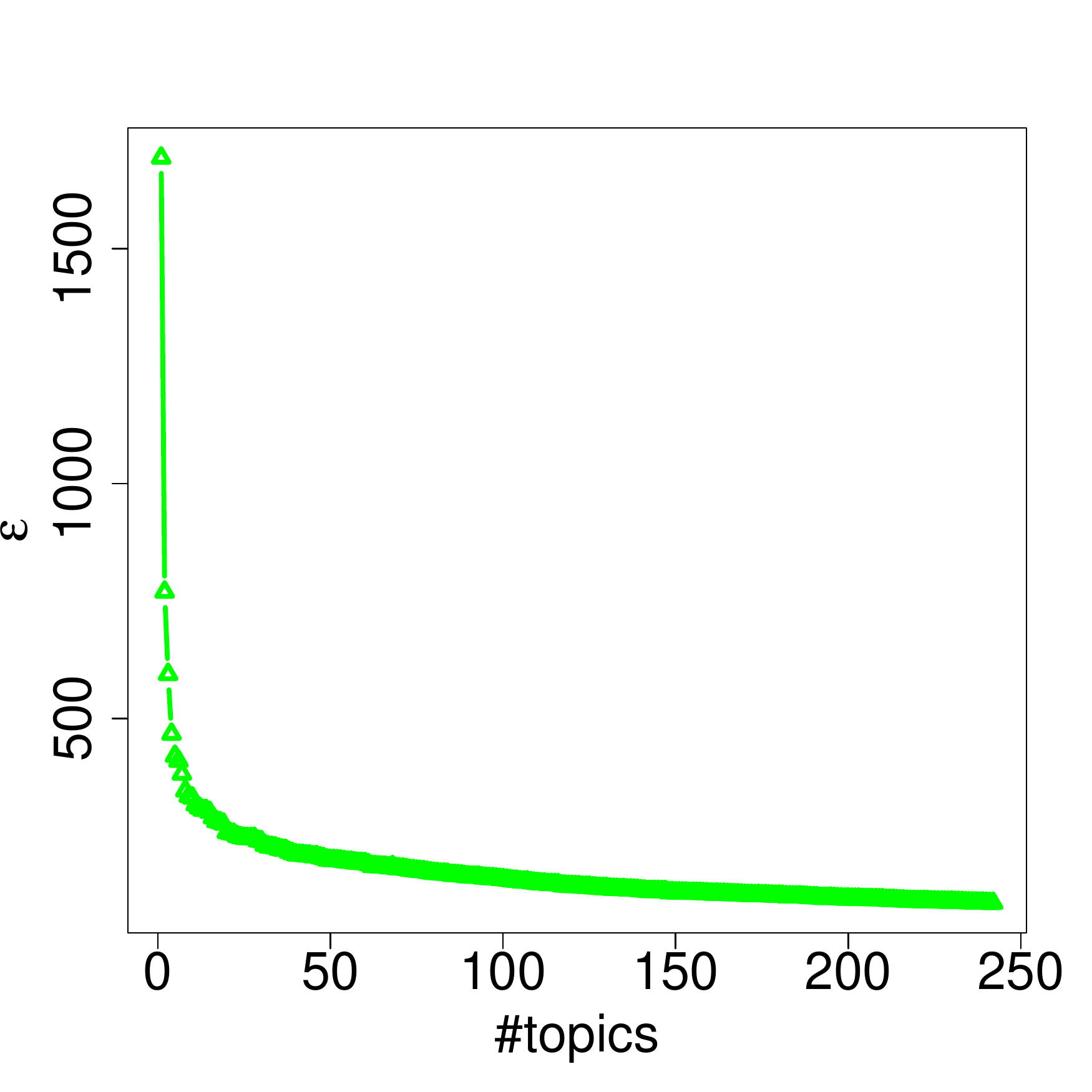}
\label{fig:subfig3}
}
\label{fig:pf-nplsa-total}
\caption{Results of the parameter-free nPLSA model on DBLP dataset in a single run.}
\end{figure}

Table 1 lists some example topics discovered by the parameter-free nPLSA, compared with the topics discovered by PLSA and LDA with the same number of topics provided. The topics discovered by nPLSA are comparable with those found by PLSA and LDA, which is promising given that the number of topics are adapted in a single run. 
 
\begin{table*}[htbp]
\caption{Example topics discovered by different models. A comparable quality is presented across models. }
\begin{center}
\scalebox{0.65}{
\begin{tabular}{|c|c|c|c||c|c|c|c||c|c|c|c|} \hline
\multicolumn{4}{|c||}{\textbf{PLSA}}&\multicolumn{4}{|c||}{\textbf{LDA}}&\multicolumn{4}{|c|}{\textbf{Parameter-free nPLSA}} \\ \hline
Topic 1& Topic 2& Topic 3&Topic 4&Topic 1&Topic 2 &Topic 3 &Topic 4& Topic 1&Topic 2&Topic 3&Topic 4\\ \hline
network& computing& classification&clustering&wireless&services &classification &real &network&service&classification&clustering\\ 
networks& services& classifier&cluster&sensor&computing&accuracy&clustering&networks&services&fuzzy&distance\\ 
nodes& grids& vector&clusters&networks&internet&vector&similarity&nodes&management&selection&measure\\ 
wireless& environments&accuracy &hierarchical&transmission&deployment &classify&cluster &sensor&resource&accuracy&similarity\\ 
sensor& resources& classifiers&cell&network&infrastructure&vectors&hierarchical&wireless&resources&decision&cluster\\ 
node& middleware& results&validity&sensing&middleware&fusion&experiments& node&middleware&method&clusters\\ 
energy& infrastructure& evolutionary&cells&sensors&paradigm &classifiers &clusters& protocol&architecture&set&measures\\ 
protocol& workflow& classify&caching&due&ubiqutous &normal&measure& transmission&support&classifier&hierarchical\\ 
paper& computational& classes&proposed&cooperative&environments &predictive &effectiveness& paper&paper&criteria&metric\\ 
transmission&heterogeneous& proposed& paper&interference&technologies &classify &synthetic& protocol&based&proposed&validity \\  \hline
\end{tabular}
}
\end{center}
\label{topic comparison of different models with the same topic num}
\end{table*}%

\begin{figure}[htbp]
\centering
\subfigure[Time vs $\#$ documents]{
\includegraphics[width=1.5in]{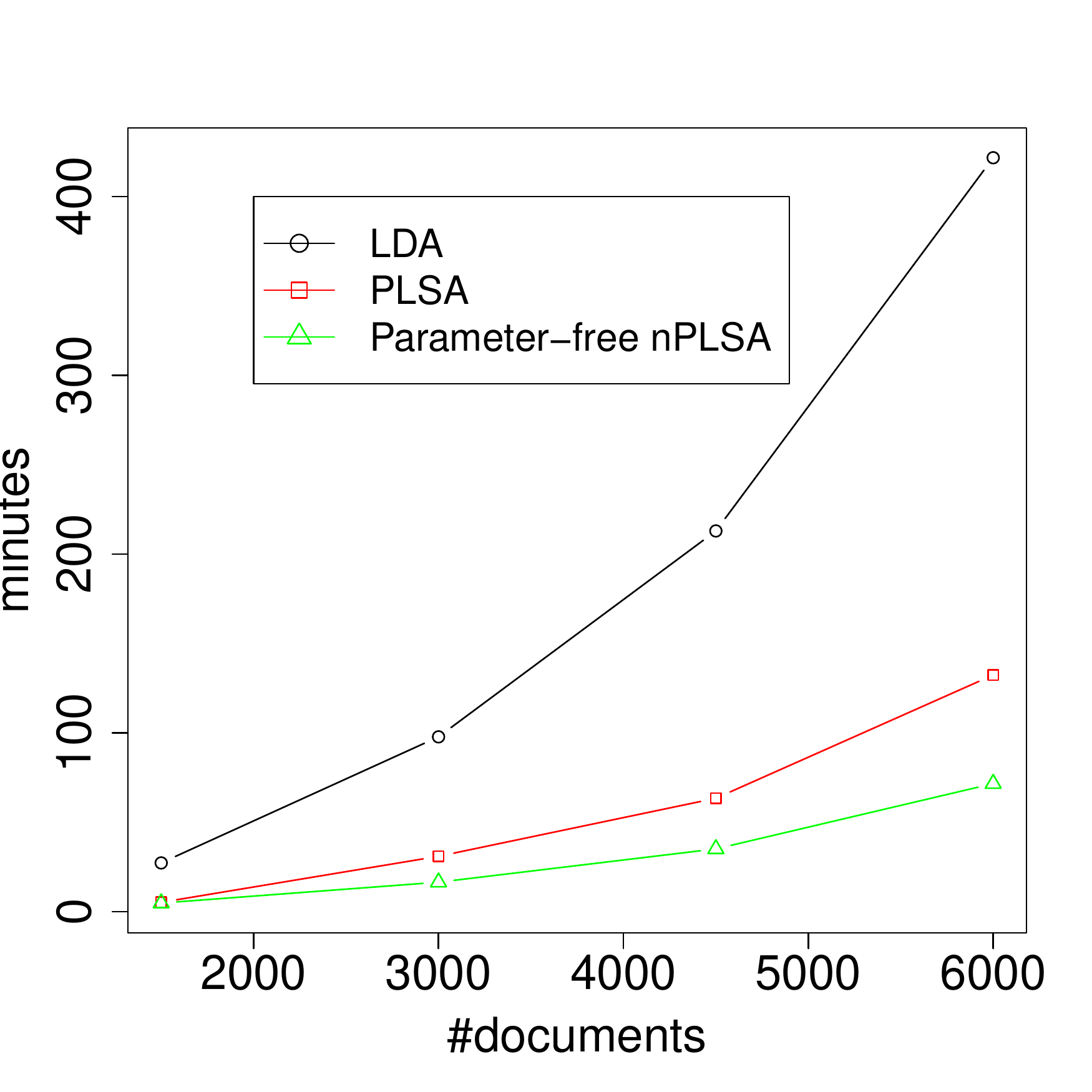}
\label{fig:time-size}
}
\subfigure[Time vs $\#$ topics]{
\includegraphics[width=1.5in]{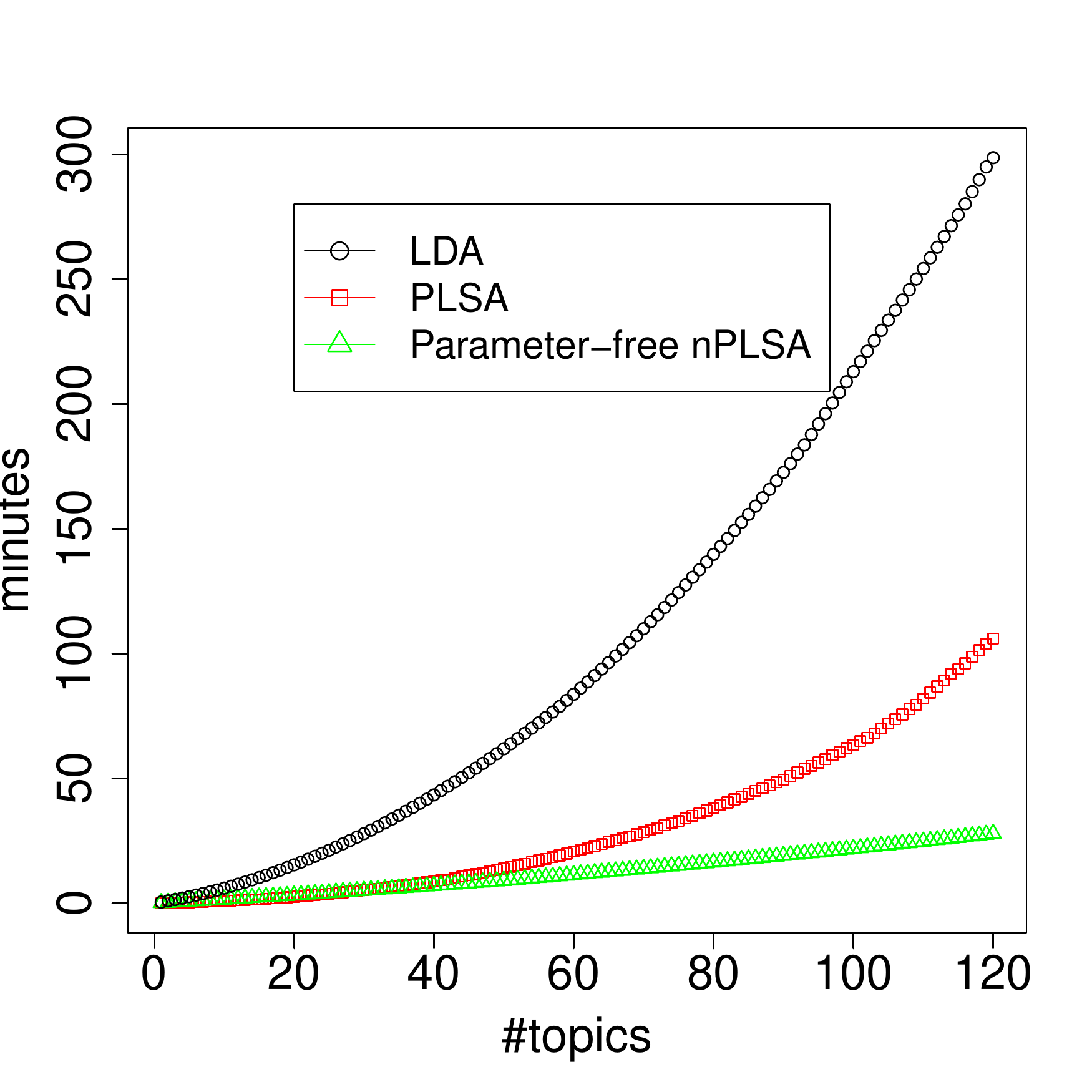}
\label{fig:time-topic}
}
\label{fig:time}
\caption{Comparison of time complexity. Compared to PLSA/LDA, the parameter-free nPLSA becomes  more efficient when there are more training documents (a) or more topics in the training documents (b). For PLSA and LDA, different runs are conducted with topic number ranging from 1 to the topic number discovered by the parameter-free nPLSA. In LDA, 500 Gibbs sampling iterations are conducted to infer the topics.}
\end{figure}

In Figure 5, we also compared the efficiency of finding the number of topics using the parameter-free nPLSA versus using enumerative runs of PLSA/LDA. The parameter-free nPLSA is significantly more efficient, especially when the size of the data increases (Figure~\ref{fig:time-size}). The time complexity of nPLSA grows linearly with the number of topics discovered, while enumerating different numbers of topics using LDA or PLSA results in an quadratic growth with the number of topics (Figure~\ref{fig:time-topic}). This suggests that the parameter-free nPLSA is especially suitable for bigdata from an open domain, where there exists lots of topics. 

\vspace{1em}\noindent \textbf{Results of the weakly-supervised nPLSA on DBLP}\vspace{1em}

\begin{figure*}[htdp!]
\centering
\subfigure[Query: ``Data Mining'']{
\includegraphics[width=3.3in]{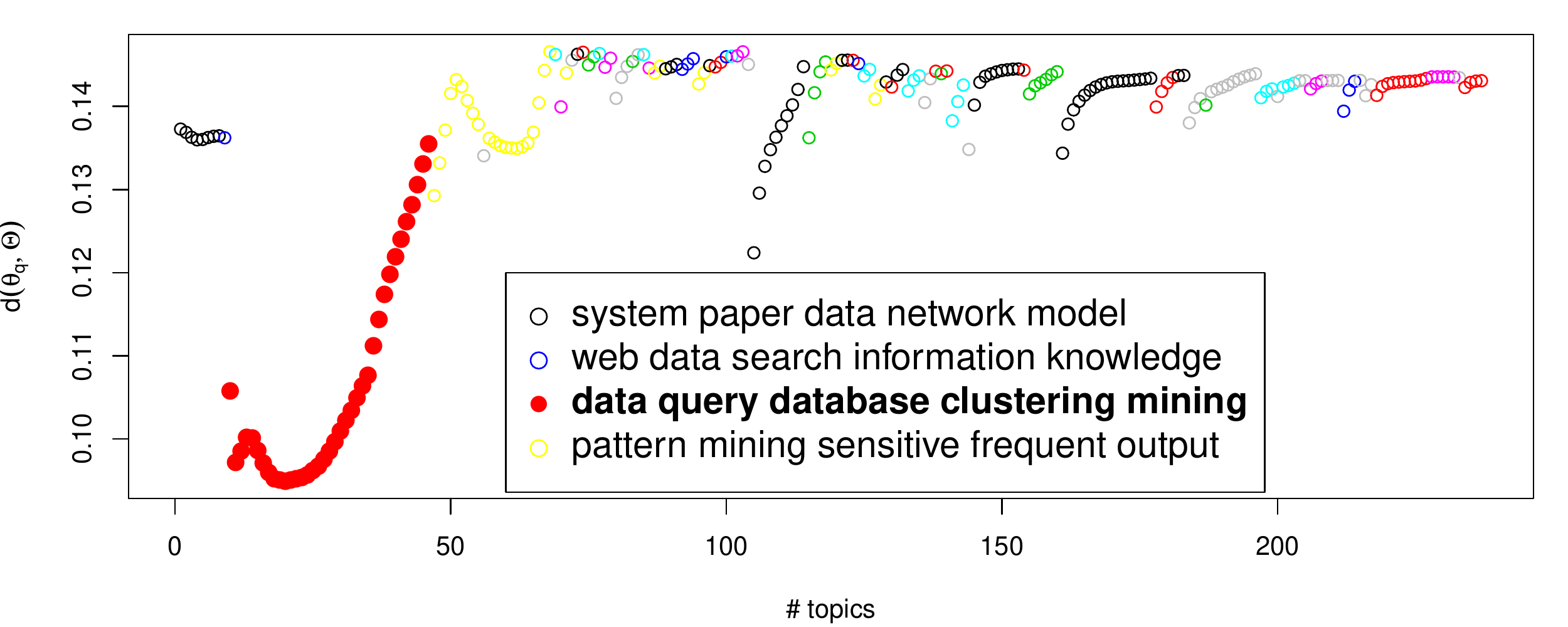}
\label{fig:DM}
}
\subfigure[Query: ``Frequent Pattern Mining'']{
\includegraphics[width=3.3in]{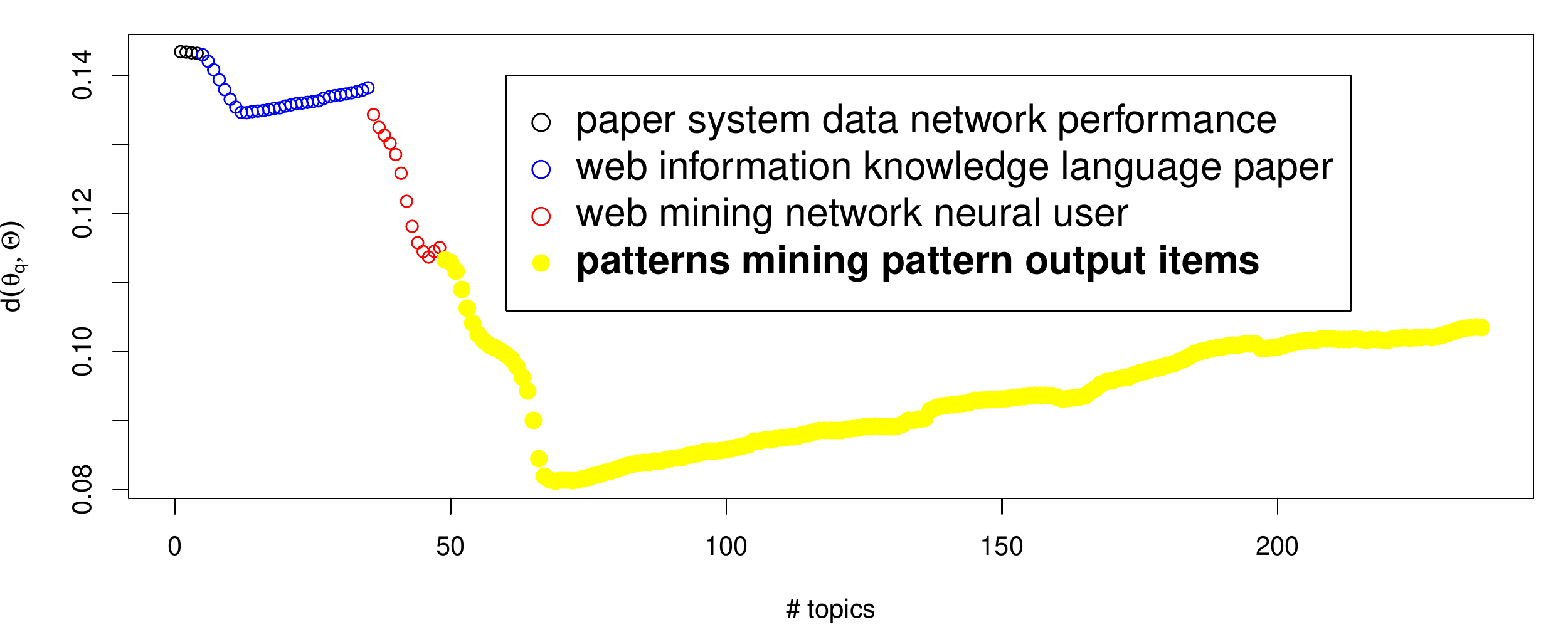}
\label{fig:FPM}

}

\caption{Examples of the weakly-supervised nPLSA model with different queries. Two figures are presented for each query. The top figure shows how the minimal distance between the query and the topics $d(\theta_q,\Theta)$ changes along the iterations (number of topics). The change of color indicates a split of this  closest topic into subtopics. The first four versions of the closest topic to the query are presented. \label{fig:query-by-example}}
\label{fig:querybyexample}
\end{figure*}

The observations above prove that the parameter-free treatment of nPLSA performs well in reality and learns the appropriate number of topics from the data. What if the user specifies a personalized preference of the granularity of topics? We move forward to evaluate the weakly-supervised nPLSA model, which allows the user to query with an exemplar topic. The model will stop generating new topics when the distance of the closest topic to the query is minimized. In Figure~\ref{fig:query-by-example}, we present the results of the weakly-supervised nPLSA model with two different queries. For each query, we plot how the distance between the closest topic and the query, $d(\theta_q, \Theta)$, changes w.r.t the number of topics discovered. Top five words of the highest probability are listed to present the meaning of the topic. Figure~\ref{fig:DM} studies the query ``data mining.'' In the first several iterations, the closest topic to this query presents words like ``\textit{system, paper, data, network, model},'' which is more likely a background topic of the corpus. After several iterations, the closest topic becomes ``\textit{web, data, search, information, knowledge},'' which seems to be a mixture of information retrieval and data mining. When more topics are generated, the closest topic becomes ``\textit{data query database clustering mining},'' which is precisely the ``data mining'' topic the user is looking for. There we can see that the distance $d(\theta_q, \Theta)$ achieves the minimum around 25 iterations. 
At this point the algorithm should stop generating new topics. However, if the algorithm continues to generate new topics, the next closest topic to the query will become ``\textit{pattern, mining, sensitive, frequent, output},'' which is about ``frequent pattern mining,'' a subtopic of ``data mining.'' In this case, the granularity of the topics becomes too small comparing to the user's preference. 
We then study another query ``frequent pattern mining,'' which indicates a finer granularity of interest than ``data mining.'' 
From Figure~\ref{fig:FPM}, we can see that distance $d(\theta_q, \Theta)$ achieves the minimum around 70 topics, indicating that there are more topics at the granularity of ``frequent pattern mining'' than those comparable to ``data mining.''  The shortest distance becomes stable after the topic ``\textit{pattern, mining, patterns, output, items}'' is found. This is precisely the ``frequent pattern mining'' topic, which does not further split along the iterations.



Table 2 list some exemplar topics discovered by the weakly-supervised nPLSA model given different exemplar queries. We can see that the granularity of those topics is close to the query topic and well satisfies users' personal needs.

\begin{table*}[htbp!]
\caption{Example topics by weakly-supervised nPLSA with different queries. Topics with a similar granularity to the query are extracted. The query ``frequent pattern mining'' yields topics with a finer granularity. }
\begin{center}
\scalebox{0.65}{
\begin{tabular}{|c|c|c|c||c|c|c|c|}\hline
\multicolumn{4}{|c||}{\textbf{Query: ``Data Mining''}}&\multicolumn{4}{|c|}{\textbf{Query: ``Frequent Pattern Mining'' (FPM)}} \\ \hline
``Data Mining''&``Information Retrieval'' &``Wireless Networks''&``Machine Learning''&``FPM''&``Security Protocols''&``Image Reconstruction''&``Distributed Computing''\\ \hline
data&information&network&learning &patterns&security&3d&distributed\\
query&web&networks&training &mining&protocols&view&communication\\
database&knowledge&nodes&approach &output&protocol&reconstruction&task\\
large&search&protocol&results  &items&secure&compact&groups\\
clustering&semantic&energy&experiments &sensitive&response&precision&tasks\\
queries&documents&wireless&methods &frequent&attacks&geometry&parallel\\
methods&texts&scheme&sequences &discovery&public&2d&message \\
techniques&rules&routing&genes &association&technical&geometric&clusters\\
index&paper&sensor&data &list&events&greatly&message\\
paper&retrieval&paper&signal &utility&attack&paper&cluster \\ \hline
\end{tabular}
}
\end{center}
\label{default}
\end{table*}

In our experiments, we also applied the nPLSA model and its parameter-free versions to a social media dataset collected from Twitter. 
As the length of tweets is too short for topic modeling analysis~\cite{Hong:2010:EST:1964858.1964870}, we adopt the same strategy as~\cite{Hong:2010:EST:1964858.1964870} by treating each hashtag as a ``pseudo document'' and concatenate all the tweets containing this hashtag into the same document. A sample of 10,769 ``documents'' are collected from a seven-days time window, with a vocabulary of 121,709 words. Similar behaviors of the models are observed on the Twitter dataset. Due to the space limit, we omit the detailed description of the results on the Twitter dataset.

	\section{Conclusion}
We presented a series of treatments to the classical topic model, PLSA, in order to eliminate the arbitrary predetermination of the number of topics. We first proposed a nonparametric topic model named nPLSA, which makes it possible to grow the number of topics in a single inference procedure that remains simple and efficient. Two parameter-free treatments of nPLSA are then presented that truly eliminate all arbitrary parameters. One of them finds the number of topics that are the most distinctive from each other, and the other allows a user to provide an exemplar topic and finds topics at the same granularity of the example. The nPLSA model and its parameter-free treatments have considerable advantage over parametric topic models such as PLSA and LDA, and Bayesian nonparametric models such as HDP. Future work will focus on further scaling these models for big data. In particular, it is possible to develop an online inference procedure of these models similar to the practice in~\cite{DBLP:conf/nips/HoffmanBB10} and~\cite{DBLP:journals/jmlr/WangPB11}, as well as distributed versions of the algorithms. We also plan to investigate alternative stopping criteria for the parameter-free inference procedure.

	\bibliographystyle{abbrv}
	\vspace{2em}
	\small{
		\bibliography{sigproc}
	}
\end{document}